\documentclass[10pt, conference, letterpaper]{IEEEtran}

\usepackage{amsfonts}       % blackboard math symbols
\usepackage{nicefrac}       % compact symbols for 1/2, etc.
\usepackage{graphicx}
\usepackage{microtype}      % microtypography
\usepackage{xcolor}         % colors
\usepackage{amsmath}
\usepackage{amssymb}
\usepackage{amsfonts}
\usepackage{amsthm}
\usepackage{verbatim}
\usepackage{algorithmic}
\usepackage{algorithm}
\usepackage{subfigure}
\allowdisplaybreaks[4]
\onecolumn
\title{DRAG: Divergence-based Adaptive Aggregation in Federated learning on Non-IID Data}

\author{Feng Zhu, Jingjing Zhang, Shengyun Liu and Xin Wang \\
School of Information Science and Technology, Fudan University\\
School of Electronic, Information and Electrical Engineering, Shanghai Jiao Tong University \\
\texttt{\{20210720072,jingjingzhang,xwang11\}@fudan.edu.cn} \\
\texttt{shengyun.liu@sjtu.edu.cn
}}

\begin{document}
% \author[1]{Feng Zhu}
% \author[1]{Jingjing Zhang\thanks{Corresponding author: jingjingzhang@fudan.edu.cn}}
% \author[2]{Shengyun Liu}
% \author[1]{Xin Wang}

% \affil[1]{School of Information Science and Technology, Fudan University}
% \affil[2]{School of Electronic, Information and Electrical Engineering, Shanghai Jiao Tong University}
\maketitle

\begin{abstract}
% Local stochastic gradient descent (local SGD) is a method aimed at improving communication efficiency in FL by allowing each worker to perform local updates. Nonetheless, under heterogeneous data distribution across working nodes, each worker will update its local model towards a local optimum, and simply averaging them will cause the convergence rate to slow down, known as the ``client-drift'' phenomenon. To deal with this issue, we propose the divergence-based adaptive aggregation (DRAG) algorithm to adaptively ``drag'' the local update of each worker in each training round according its ``degree of divergence'' through vector manipulation. Rigorous convergence result of DRAG is established. Also, numerical results demonstrate that the proposed DRAG method outperforms state-of-the-art algorithms in coping with client-drift.
% State-of-the-art algorithms such as SCAFFOLD and AdaBest utilize control variates to mitigate this effect, but causing increased communication overhead and unsteady performance. 
Local stochastic gradient descent (SGD) is a fundamental approach in achieving communication efficiency in Federated Learning (FL) by allowing individual workers to perform local updates. However, the presence of heterogeneous data distributions across working nodes causes each worker to update its local model towards a local optimum, leading to the phenomenon known as ``client-drift" and resulting in slowed convergence. To address this issue, previous works have explored methods that either introduce communication overhead or suffer from unsteady performance. In this work, we introduce a novel metric called ``degree of divergence," quantifying the angle between the local gradient and the global reference direction. Leveraging this metric, we propose the divergence-based adaptive aggregation (DRAG) algorithm, which dynamically ``drags" the received local updates toward the reference direction in each round without requiring extra communication overhead. Furthermore, we establish a rigorous convergence analysis for DRAG, proving its ability to achieve a sublinear convergence rate. Compelling experimental results are presented to illustrate DRAG's superior performance compared to state-of-the-art algorithms in effectively managing the client-drift phenomenon. Additionally, DRAG exhibits remarkable resilience against certain Byzantine attacks. By securely sharing a small sample of the client's data with the FL server, DRAG effectively counters these attacks, as demonstrated through comprehensive experiments.
\end{abstract}


\section{Introduction}
% As the complexity of machine learning tasks increasingly grows and the volume of data also multiplies at an exponential rate, distributed implementation of working nodes, i.e., federated learning (FL), has gained rising attention \cite{dean2012large,smith2017federated,lim2020federated,yang2019federated}. Parameter-server (PS) setting is one of the most widely employed paradigms in FL, where the PS broadcasts the latest global model to the workers for computation and the workers send their computed local models back to the PS for aggregation and update \cite{lian2017can,li2014scaling,gupta2016model}. By developing a novel method named divergence-based adaptive aggregation (DRAG), this paper aims to deal with the following two issues. In this work, we address the challenges posed by the growing complexity and scale of machine learning tasks, with a specific focus on FL involving heterogeneous data. Our primary aim is to overcome the following two critical challenges:

With the increasing complexity of machine learning tasks and the exponential growth in data volume, the adoption of distributed implementations, e.g., federated learning (FL), has been gaining considerable attention \cite{dean2012large,smith2017federated,lim2020federated,yang2019federated}. The parameter-server (PS) setting stands as one of the most widely utilized paradigms in FL. In this approach, the PS broadcasts the latest global model to the workers for computation, while the workers, in turn, send their computed local models back to the PS for aggregation and update \cite{lian2017can,li2014scaling,gupta2016model}. 

Due to the frequent bidirectional transmissions between the PS and the workers, communication efficiency has become a critical bottleneck in large-scale FL \cite{zhang2015deep, alistarh2017qsgd}.  To overcome this challenge, the local stochastic gradient descent (SGD) method \cite{zinkevich2010parallelized,mcmahan2017communication} has been introduced. This method enables each worker to perform multiple local updates before uploading the latest model, significantly enhancing communication efficiency.
% The authors of \cite{zinkevich2010parallelized} first put forth the idea of local SGD to improve communication efficiency. Specifically, each client optimizes its own local model and the PS aggregates the latest local models as the final output. Nevertheless, this method is proved to perform worse than distributed SGD under worst-case scenarios, particularly when dealing with non-convex or non-smooth objective functions \cite{arjevani2015communication,zhang2012communication}. In response to this limitation, 
\cite{mcmahan2017communication} proposes the widely-studied federated averaging (FedAvg) algorithm, where the communication frequency between the PS and the workers is increased, leading to improved performance. Under heterogeneous data distribution, FedAvg achieves remarkable performance compared to parallel SGD which does not employ local updates, which is also theoretically substantiated by rigorous convergence analysis \cite{zhou2018convergence,woodworth2018graph,wang2019cooperative,khaled2020tighter}. 

This paper focuses on FL with local SGD involving heterogeneous data. In particular, it aims to address the following two challenges:

% This paper introduces a novel method called divergence-based adaptive aggregation (DRAG) to address the challenges presented by the growing complexity and scale of machine learning tasks, with a particular focus on FL involving heterogeneous data. The primary aim of DRAG is to overcome the following two critical challenges:

\textbf{Challenge 1: Client-Drift.} The phenomenon of ``client-drift" in local SGD is first identified in \cite{zhao2018federated}. The authors observe that the local models of different workers tend to converge to local optima when dealing with heterogeneously distributed data. Consequently, the straightforward averaging of these local models results in poor convergence outcomes. State-of-the-art algorithms such as SCAFFOLD \cite{karimireddy2020scaffold} and AdaBest \cite{varno2022adabest} that utilize control variates to mitigate client-drift usually have unsteady performance and cannot adapt to diverse settings.

% First discovered in \cite{zhao2018federated}, the authors identify the phenomenon of ``client-drift'' in local SGD, namely that the local models of different workers will converge to local optima under heterogeneously distributed data, and the simple direct averaging of the local models will cause terrible convergence results.

To address this issue, we propose a method named divergence-based adaptive aggregation (DRAG). The proposed DRAG algorithm introduces a novel metric named ``degree of divergence'', which quantifies the extent of the local update of each worker in each round diverging from the reference direction. The reference direction, having a momentum form, is a weighted sum of all the historical global update directions. Leveraging the newly defined metric, the local update of each worker is then dynamically ``dragged'' toward the reference direction through weighted vector manipulation. The PS finally aggregates the dragged local updates to update the global model. This approach can effectively mitigate the client-drift phenomenon while preserving the diversity of local gradients and accelerates the convergence process in distributed learning scenarios with heterogeneous data distributions.

\textbf{Challenge 2: Byzantine attack.} As FL operates in a distributed manner, it is susceptible to adversarial attacks launched by malicious clients, commonly known as Byzantine attacks \cite{so2020byzantine,cao2020fltrust,bagdasaryan2020backdoor}. For instance, one type of Byzantine attack involves the malicious client reversing the direction of the local update or scaling the local gradient by a factor to negatively influence the training process \cite{prakash2020mitigating}. In this work, the proposed DRAG method is proven to be effective in mitigating this type of attack through vector manipulation, ensuring the integrity and security of the FL process even in the presence of malicious clients.

% \subsection{Our Contributions}
\textbf{Our contributions:} Motivated by these challenges, we investigate a PS-based framework under heterogeneous data distribution. The main contribution can be summarized as follows. 
\begin{itemize}
    \item We introduce a new metric called ``degree of divergence" that utilizes the angle to quantify the deviation between workers' local updates and the weighted sum of historical global updates. This metric forms the basis for our proposed method, divergence-based adaptive aggregation (DRAG), effectively addressing the client-drift phenomenon and accelerating the convergence rate.  
    \item We establish a rigorous convergence analysis of the DRAG algorithm, demonstrating that it achieves a sublinear convergence rate similar to other local SGD methods.
    \item The performance of DRAG is evaluated on EMNIST and CIFAR-10 datasets, and we conduct a comprehensive comparison with state-of-the-art algorithms, revealing that our method consistently achieves superior results.
    \item Finally, we demonstrate that DRAG exhibits resilience against certain Byzantine attacks, such as reversing the local update direction or scaling the local update.
\end{itemize}

\section{Related Works}
% In this section, we briefly review previous works on local SGD and on dealing with the client-drift issue caused by data heterogeneity. After that, we also introduce some background information regarding byzantine attacks.
In this section, we provide a brief review of prior research on strategies employed to address the client-drift issue resulting from data heterogeneity. Furthermore, we offer background information on byzantine attacks.

% As the performance of local SGD is tested on non-convex objective functions, it is straightforward to test it on non-i.i.d. data, which is when the client-drift problem is identified \cite{zhao2018federated}. The authors find that the accuracy reduces significantly with each client training a single class of data, and they propose to improve the performance by assigning a shared subset of data to all the clients. The main cause of this phenomenon is that the local optima of the workers diverge too far from the aggregated update direction. 
\cite{zhao2018federated} initiates a series of research attempting to deal with the so-called ``client-drift'' issue.
% Additionally, the convergence rate of FedAvg is analyzed under non-i.i.d. data as well, by assuming bounded gradient variance \cite{li2019convergence, haddadpour2019local,wang2019adaptive2,yu2019parallel,yang2020achieving}.
There are several categories of methods used to address client-drift in FL.
One is to incorporate the idea of variance reduction \cite{defazio2014saga,johnson2013accelerating} into local SGD such as \cite{liang2019variance,li2019feddane,pathak2020fedsplit,konevcny2016federated}. However, these methods often require the full participation of workers, making them less practical when only a subset of devices are active.

Another category involves using control variates to compensate for the drift. \cite{karimireddy2020scaffold} proposes SCAFFOLD that uses local and global control variates to correct the drift on the client side. Building on SCAFFOLD, \cite{karimireddy2020mime} further adopts the idea of momentum to for server-level optimization. Additionally, employing local and global variates, FedDyn  \cite{acar2021federated} and AdaBest \cite{varno2022adabest} correct the drift on both the server and client sides where Adabest did not use historical information for the global variate, unlike FedDyn.

Another effective approach to reduce client-drift is through explicit gradient constraint. FedProx, introduced in \cite{li2020federated}, adds a regularization term in the objective function to prevent the drift from being too far. Decoupling the local and global model, \cite{gao2022feddc} utilizes both control variates and regularization terms to jointly learn the gap between the local model and the global one.

% In contrary to all the methods mentioned above, the proposed DRAG is a heuristic and intuitive method that drags each local model towards the reference direction according to the extent of the drift via vector manipulation. The details of the method is elaborated in Section 4.

% Next, some background knowledge about byzantine attacks is reviewed. Due to the distributed manner of federated learning, it is vulnerable to attacks once some of the clients experience some software bugs or are even hacked by people with malicious intentions, which is widely known as the byzantine attacks \cite{so2020byzantine,cao2020fltrust,prakash2020mitigating}.

Next, we provide a brief overview of byzantine attacks in the context of FL. Due to the distributed nature of FL, it becomes vulnerable to attacks if some of the clients experience software bugs or fall under the control of malicious entities. These attacks, widely known as byzantine attacks \cite{so2020byzantine,cao2020fltrust,prakash2020mitigating}, can potentially compromise the integrity of the learning process and the security of the overall system. It is essential to develop defense mechanisms to counter such attacks and ensure the robustness and reliability of federated learning in the presence of adversarial clients.

Apart from the reversing direction attack mentioned in the introduction part, there are other more advanced attacks in the literature, which we briefly introduce as follows. The Label flipping attack is simple to implement: In supervised learning, the malicious client flips the labels of its training examples in a reverse way, or any other way that induces a mismatch between the data example and its label \cite{fang2020local}. Krum attack \cite{blanchard2017machine} and Trim attack \cite{fang2020local} are coupled with their corresponding gradient aggregation rules.

% To be specific, these two attack modes try to solve an optimization problem for particular aggregation rules (Krum or Trim) that aims to change the correct global update direction the most through poisoning the local model updates of the malicious clients. The malicious client that initiates a Scaling attack trains a backdoor-poisoned dataset and scales the byzantine local model update by multiplying it with a factor much larger than 1. The reason for doing that is to magnify the influence of the poisoned local update and prevent it from being diluted via averaging.

One of the approaches to defending against byzantine attacks is to remove the outliers in the clients through utilizing byzantine-robust aggregation rules, i.e., comparing the local updates of the clients and getting rid of the anomalies \cite{blanchard2017machine,fang2020local,yin2018byzantine}. For instance, the Krum aggregation rule \cite{blanchard2017machine} uses the one local update that has the smallest Euclidean distance to the rest of the clients to update the global model. However, this kind of method loses its robustness against byzantine attacks when malicious clients take up a large proportion since it lacks \textit{root of trust}. To tackle this problem, \cite{cao2020fltrust} proposes the FLTrust algorithm where the server maintains a small subset of the correct dataset named the ``\textit{root dataset}''. By doing this, the server then achieves a trusted source of each global update direction and improves its robustness against attacks.

The rest of the paper is organized as follows. Section III describes the system model. The development of the proposed DRAG scheme is delineated in Section IV. Section V presents the convergence analysis of DRAG. Numerical results are provided in Section VI. Section VII concludes the work.

\section{System Model}

In this paper, we investigate an FL architecture consisting of $M$ workers denoted by the set $\mathcal{M}:=\{1,..., M\}$. Each worker $m$ maintains a local dataset $\mathcal{D}_m$ with a size of $N_m$. These datasets are drawn from a global dataset $\mathcal{D}=\{z_i\}_{i=1}^N$,i.e., we have $\mathcal{D}=\bigcup_{m\in\mathcal{M}} \mathcal{D}_m$. The objective is to solve the minimization problem, where the objective function is defined as the average sum of local functions contributed by individual workers:
\begin{align}
    \min_{\boldsymbol{\theta}\in \mathbb{R}^d} f(\boldsymbol{\theta})&=\frac{1}{M}\sum_{m\in\mathcal{M}}F_m(\boldsymbol{\theta}),\nonumber \\ \text{with}\quad F_m(\boldsymbol{\theta})&:=\mathbb{E}\left[F_m(\boldsymbol{\theta}; z_m)\right], m\in\mathcal{M},\label{problem}
\end{align}
where parameter $\boldsymbol{\theta}$ with dimension $d$ is the variable to be optimized and $\{F_m(\boldsymbol{\theta}),m\in\mathcal{M}\}$ are smooth functions, and $z_m$ is a sample randomly selected from the local dataset $\mathcal{D}_m$ of worker $m$.

With the objective function defined in (\ref{problem}), we attempt to solve the problem in an iterative manner with local SGD. In particular, at each training round $j$, the PS broadcasts the latest global model $\boldsymbol{\theta}^t$ to a subset $\mathcal{S}^t$ of 
 $S$ randomly selected workers. Then each worker $m$ sets its local model $\boldsymbol{\theta}_m^{t,0}=\boldsymbol{\theta}^t$ and performs $U$ local updates via the following formula
\begin{equation}
\boldsymbol{\theta}_{m}^{t,u+1}=\boldsymbol{\theta}_{m}^{t,u}-\frac{\eta}{B} \sum_{b=1}^B \nabla F(\boldsymbol{\theta}_{m}^{t,u}; z_{m,b}^{t,u}), \label{update}
\end{equation}
for local iteration $u=0,...,U-1$. Note that $\eta$ is the stepsize; $B$ is the mini-batch size; $\frac{1}{B} \sum_{b=1}^B \nabla F(\boldsymbol{\theta}_{m}^{t,u}; z_{m,b}^{t,u})$ is the mini-batch gradient to be computed by worker $m$ at iteration $u$ and $z_{m,b}^{t,u}$ are drawn independently from dataset $\mathcal{D}_m$ across all workers, batches, local iterations and training rounds. Each worker $m$ then sends $\mathbf{g}_m^t$ to the server. The variable $\mathbf{g}_m^t$ represents the discrepancy between the latest local model after $U$ local updates and the original global model received at the beginning of the training round $t$. Specifically, for each worker $m$ in training round $t$, we define $\mathbf{g}_m^t$ as 
\begin{align}
 \mathbf{g}_m^t = \boldsymbol{\theta}_m^{t, U} - \boldsymbol{\theta}^{t}. 
\end{align}

Lastly, the PS aggregates the local models to update the global model:
\begin{align}
    \boldsymbol{\theta}^{t+1}=\boldsymbol{\theta}^{t}+\frac{1}{S}\sum_{m\in\mathcal{S}^t}\mathbf{g}_m^t.\label{aggregation}
\end{align}
The training process continues some convergence criterion is satisfied. 
% The setup described above is the basic procedure of FL using local SGD. The modifications to the framework of the proposed algorithm are detailed in the next section.

% The above framework is without a byzantine attack. While in the presence of byzantine attacks, In the randomly selected subset $\mathcal{S}^t$, we assume that there are $A$ malicious clients in a subset $\mathcal{A}^t$. At each training round $t$, after finishing $U$ local updates, each malicious client $m\in\mathcal{A}^t$ multiplies the local model update $\mathbf{g}_m^t$ by a scalar $p_m^t$ which can be either positive or negative and sends it to the server, i.e., reversing the update direction or scaling the module of the update. 

\begin{figure}[t!]
\centering
\includegraphics[width=3in]{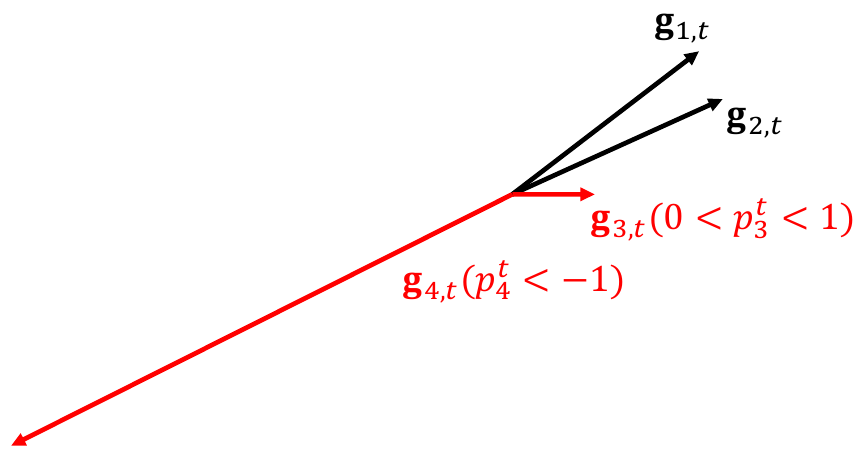}
\caption{Illustration of byzantine attacks dealt with in this work, where the black color represents the normal clients and the red color represents the malicious ones.}
\label{attack}
\end{figure}

We proceed to elaborate on the scenario with the byzantine attack that we aim to defend against in this work. In the randomly selected subset $\mathcal{S}^t$, we assume the presence of $A$ malicious clients denoted by $\mathcal{A}^t \in \mathcal{S}^t$. As shown in Fig.~\ref{attack}, during each training round $t$, after completing $U$ local updates, each malicious client $m\in\mathcal{A}^t$ manipulates their local model update $\boldsymbol{\theta}_m^{t, U} - \boldsymbol{\theta}^{t}$ by multiplying it with a scalar $p_m^t$, denote as $\hat{\mathbf{g}}_m^t=p_m^t(\boldsymbol{\theta}_m^{t, U} - \boldsymbol{\theta}^{t})$. Note that $p_m^t$ can be either positive or negative. Subsequently, the modified update $\hat{\mathbf{g}}_m^t$ is sent to the server, effectively reversing the update direction or scaling the magnitude of the update. The PS then aggregates the received local models as follows:
Lastly, the PS aggregates the local models to update the global model:
\begin{align}
    \boldsymbol{\theta}^{t+1}=\boldsymbol{\theta}^{t}+\frac{1}{S}\big(\sum_{m\in\mathcal{A}^t}\hat{\mathbf{g}}_m^t+\sum_{m\in\mathcal{S}^t \setminus \mathcal{A}^t}\mathbf{g}_m^t\big).\label{aggregation}
\end{align}

% Now we continue to elaborate on the byzantine attack that we aim to defend against in this work. In the randomly selected subset $\mathcal{S}^t$, we assume that there are $A$ malicious clients in a subset $\mathcal{A}^t$. At each training round $t$, after finishing $U$ local updates, each malicious client $m\in\mathcal{A}^t$ multiplies the local model update $\mathbf{g}_m^t$ by a scalar $p_m^t$ which can be either positive or negative and sends it to the server, i.e., reversing the update direction or scaling the module of the update. 

\section{divergence-based adaptive aggregation (DRAG) Algorithm}

In this section, we introduce the proposed divergence-based adaptive aggregation (DRAG) method. In contrast to most state-of-the-art algorithms designed to tackle  client drift by employing control variates for local and global model alignment, the proposed DRAG method adopts a heuristic and intuitive manner that ``drags" each local model toward the reference direction through vector manipulation, with the extent of dragging determined by a metric we define as the ``degree of divergence". To establish the theoretical feasibility of DRAG, we conduct a rigorous convergence analysis that provides compelling evidence of its convergence properties. 

% \subsection{Rationale behind DRAG}
\subsection{Definitions}
% In this subsection, we elaborate on the rationale behind the ideas of DRAG, including the reference direction, the degree of divergence and the vector manipulation process.

To commence, we define two crucial new variables: the reference direction and the degree of divergence. These variables play a key role in the process of dragging the local gradient.

\subsubsection{Reference Direction}
% The aim of the reference direction is to provide a feasible and reasonable direction when modifying the local gradients and to better forge a global update direction. Recall that the updating formula of the reference direction is:
The objective of the reference direction $\mathbf{r}^t$ is to offer a practical and sensible direction for modifying the local gradients, thereby facilitating the formation of an enhanced global update direction. Particularly, the updating formula for the reference direction is given as:
% \begin{align}
%     &\mathbf{r}^t=(1-\alpha)\mathbf{r}^{t-1}+\alpha\Delta^{t-1}, for\:t\geq 1\notag\\
%     &\mathbf{r}^t=\frac{1}{S}\sum_{m\in\mathcal{S}^t}\mathbf{g}_m^t, for\:t=0.\label{reference}
% \end{align}
\begin{equation}
\mathbf{r}^t=
    \begin{cases}
      (1-\alpha)\mathbf{r}^{t-1}+\alpha\Delta^{t-1}, ~~\text{for  $t \geq 1$} \\
      \frac{1}{S}\sum_{m\in\mathcal{S}^t}\mathbf{g}_m^t,~~~~~~~~~~~ \text{for  $t=0$}.
    \end{cases}\label{reference}
\end{equation}
To rewrite it as a closed-form expression, we have:
\begin{align}
    \mathbf{r}^t=\frac{(1-\alpha)^t}{S}\sum_{m\in\mathcal{S}^t}\mathbf{g}_m^t+\sum_{i=0}^{t-1}\alpha(1-\alpha)^{t-i-1}\Delta^i, ~~~t\geq 1. \label{reference_cf}
\end{align}
Here $\alpha\in(0,1)$ is some constant to control the weights of the historical directions. Furthermore,  $\Delta^t$ is the aggregated modified gradients at the PS and can be calculated as 
\begin{align}
    \Delta^t = \frac{1}{S}\sum_{m\in\mathcal{S}^t}\mathbf{v}_m^t,
\end{align}
with $\mathbf{v}_m^t$ being the modified local gradient defined below.

% It is obvious from the expression that the reference direction is a weighted sum of all the historical global update directions. Also, the weights of the most recent global updates increase as $\alpha$ becomes larger. And when $\alpha=1$, the reference direction is tantamount to the previous global gradient $\Delta^{t-1}$. Note that $\alpha$ is a tunable hyper-parameter to adaptively fit different practical scenarios.

The expression clearly demonstrates that the reference direction has a momentum form which is a weighted sum of all the historical global gradients $\Delta^i$, for $i=0,1,...,t$. Significantly, the weights assigned to the most recent global updates progressively increase as $\alpha$ grows larger. When $\alpha$ is set to 1, the reference direction becomes equivalent to the previous global gradient $\Delta^{t-1}$. Notably, the hyper-parameter $\alpha$ can be adjusted to fit different practical scenarios.

\subsubsection{Degree of Divergence}
The degree of divergence represents a fundamental aspect of the DRAG algorithm. This metric serves a critical role in measuring the extent to which the local update $\mathbf{g}_m^t$ of each worker $m$ in each training round $t$ diverges from the reference direction $\mathbf{r}^t$.

The degree of divergence is quantified by utilizing the angle $\angle_{m}^t$ between the local gradient $\mathbf{g}_m^t$ and the reference direction $\mathbf{r}^t$, given as 
\begin{align}
    \angle_{m}^t=\arccos{\frac{\left\langle\mathbf{g}_m^t, \mathbf{r}^t\right\rangle}{\|\mathbf{g}_m^t\|\|\mathbf{r}^t\|}}.
\end{align}
 By dividing $\angle_{m}^t$ over $\pi/2$, and approximating the $\arccos$ function with a linear function $y=-\frac{\pi}{2}x+\frac{\pi}{2}$ (this is one of many choices, one may also use $y=-x+\pi/2$), we define the metric of degree of divergence $\lambda_m^t$ as 
\begin{align}
    \lambda_m^t=:c\left(1-\frac{\left\langle\mathbf{g}_m^t, \mathbf{r}^t\right\rangle}{\|\mathbf{g}_m^t\|\|\mathbf{r}^t\|}\right)\in[0,2c],\label{dod}
\end{align}
where the constant $c\in[0,1]$ is a hyper-parameter, providing the flexibility to manually adjust the metric to suit various settings. It is worth noting that $\lambda_m^t$ is adaptable for each worker $m$ at each round $t$, with a larger value of $\lambda_m^t$ indicating a more significant divergence between the local gradient $\mathbf{g}_m^t$ and the reference direction $\mathbf{r}^t$.

\subsubsection{Vector Manipulation}
Utilizing the predefined reference direction $\mathbf{r}^t$ and the degree of divergence $\lambda_m^t$, we propose to drag each local gradient $\mathbf{g}_m^t$ towards the reference direction based on its degree of divergence through vector manipulation. This process yields the modified local gradient $\mathbf{v}_m^t$, given as
\begin{align}
    \mathbf{v}_m^t=(1-\lambda_m^t)\mathbf{g}_m^t+\frac{\lambda_m^t\|\mathbf{g}_m^t\|}{\|\mathbf{r}^t\|}\mathbf{r}^t.\label{vm}
\end{align}
Note that the modified gradient $\mathbf{v}_m^t$ is a weighted sum of the original local gradient $\mathbf{g}_m^t$ and the normalized reference direction $\frac{\|\mathbf{g}_m^t\|}{\|\mathbf{r}^t\|}\mathbf{r}^t$, with the weights being $1-\lambda_m^t$ and $\lambda_m^t$, respectively. By adaptively tuning the hyper-parameter $c$, we can effectively reduce the client drift while preserving the diversity of the local gradients. Additionally, the reference direction $\mathbf{r}^t$ is normalized to match the norm of $\mathbf{g}_m^t$, ensuring that the modified gradient $\mathbf{v}_m^t$ consistently has a greater component aligned with $\mathbf{r}^t$ compared to the original gradient $\mathbf{g}_m^t$. 

\textbf{Remark 1.} 
While the existing methods \cite{li2020federated} generally utilize the norms of the local gradients to quantify the degree of similarity of the local functions, they may not be suitable for handling heterogeneous data distributions, as the local gradients often diverge from the global gradient. Thus, we propose a more appropriate and reasonable metric to quantify the dissimilarity in such scenarios.

\begin{figure}[t!]
\centering
\subfigure[$0<\lambda_m^t\leq 1$]{\includegraphics[width=1.8in]{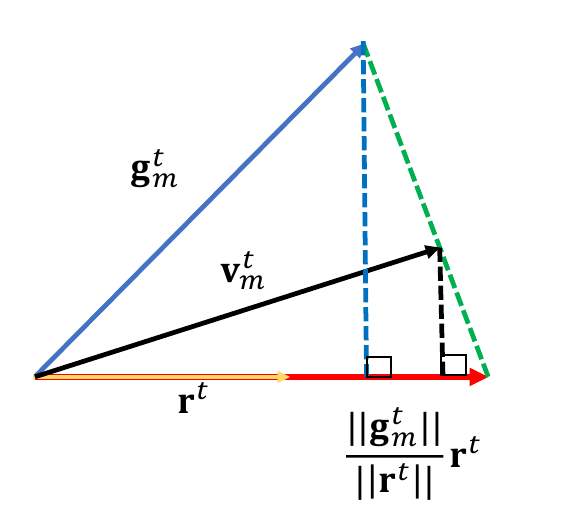}}
\subfigure[$1<\lambda_m^t\leq 2$]{\includegraphics[width=2.5in]{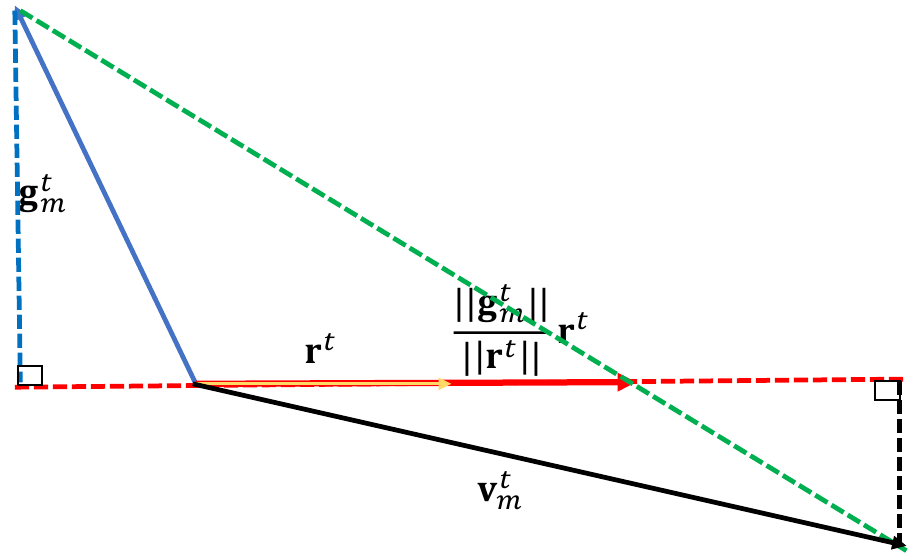}}
\caption{Illustration of vector manipulation of DRAG. It is clear that the modified gradient $\mathbf{v}_m^t$ (solid black line) has a larger component on the reference direction $\mathbf{r}^{t}$ (solid yellow line) than the original one $\mathbf{g}_{m}^t$ (solid blue line).}
\label{vectorm}
\end{figure}

\textbf{Remark 2.} 
% With the vector manipulation operation, if $0<\lambda_m^t\leq 1$, the resulted $\mathbf{v}_m^t$ both mitigates the drift and at the same time preserves the diversity of each local gradient; if $1<\lambda_m^t\leq 2$, indicating that the local gradient diverges too much from the reference direction, the $\mathbf{g}_m^t$ portion will be reversed according to (\ref{vm}) so as to abide by the correct update direction.
An illustration demonstrating how the degree of divergence $\lambda_m^t$ and the reference direction $\mathbf{r}^t$ can be utilized to guide the local gradient is provided in Fig.~\ref{vectorm}. Specifically, using the vector manipulation operation, when $0<\lambda_m^t\leq 1$, the resulting $\mathbf{v}_m^t$ effectively mitigates the drift while preserving the diversity of each local gradient. Conversely, when $1<\lambda_m^t\leq 2$, indicating that the local gradient diverges in the opposite direction to the reference direction, the $\mathbf{g}_m^t$ component is reversed according to (\ref{vm}) to ensure adherence to the correct update direction.

\subsection{Algorithm Description}
With the above definitions, we proceed to present the details of the proposed algorithm, consisting of three steps in each training round.

\textbf{Step 1:} At each training round $t$, the PS first broadcasts the current global parameter $\boldsymbol{\theta}^{t}$ to the workers in a random selected set $\mathcal{S}^t$. 

\textbf{Step 2:} Each worker in $\mathcal{S}^t$ then performs $U$ local updates via (\ref{update}) and sends its the difference $\mathbf{g}_m^t= \boldsymbol{\theta}_m^{t, U} - \boldsymbol{\theta}^{t}$ back to the PS.

\textbf{Step 3:} The PS first calculates the reference direction $\mathbf{r}^t$ via (\ref{reference_cf}) and the degree of divergence $\lambda_m^t$ via (\ref{dod}). With these two variables, the PS then drags each $\mathbf{g}_m^t$ towards the reference direction $\mathbf{r}^t$ via \eqref{vm}, yielding the modified gradient $\mathbf{v}_m^t$. In the end, the PS aggregates the modified gradients with $\Delta^t=:\frac{1}{S}\sum_{m\in\mathcal{S}^t}\mathbf{v}_m^t$ and updates the global model with $\boldsymbol{\theta}^{t+1}=\boldsymbol{\theta}^t-\Delta^t$.

\textbf{Usefulness of the degree of divergence.}
% From Figure 1 we can observe that the modified gradient $\mathbf{v}_m^t$ has reduced client drift and at the same time preserved its gradient diversity since it has more component on the reference direction and preserves the the component on its own direction.
% To briefly sum up the DRAG algorithm, unlike most control-variates-based algorithms that compensate for the drift, the proposed DRAG method has an intuitive geometric explanation. According to each worker's degree of divergence, the local gradient is adaptively dragged towards the reference direction, which has a momentum form and provides an insight for the desired update direction. The vector manipulation both deals with the client drift issue and prevents the local gradients from being over-corrected by preserving its gradient diversity. The reference direction and the degree of divergence both incorporates hyper-parameters to adapt themselves to different practical scenarios.
As illustrated in Fig.~\ref{vectorm}, it can be seen that the modified gradient $\mathbf{v}_m^t$ substantially reduces client drift while concurrently preserving the diversity of gradients. This is achieved by increasing the component aligned with the reference direction while maintaining the component in the original direction.

To summarize the DRAG algorithm, it stands apart from control-variates-based methods by offering an intuitive geometric explanation. By adaptively dragging the local gradient towards the reference direction based on each worker's degree of divergence, the algorithm incorporates momentum and provides valuable insights for the desired update direction. The vector manipulation effectively tackles the client drift issue while preventing excessive correction of local gradients, thereby preserving their diversity. Both the reference direction and the degree of divergence include hyper-parameters that enable adaptation to different practical scenarios.

\subsection{Defending Against byzantine Attacks}
% In this subsection, we demonstrate the necessary modifications made to the DRAG algorithm to render it robust to byzantine attacks, i.e., the malicious clients can reverse the gradient direction or scale the module of it

In this subsection, we illustrate the essential adaptations made to the DRAG algorithm to ensure its robustness against byzantine attacks. These attacks involve malicious clients attempting to reverse the gradient direction or scale its module. By incorporating specific modifications, DRAG is fortified to handle such adversarial behavior while maintaining the accuracy and security of the federated learning process.

 The malicious effects of attackers can undermine the usefulness of the reference direction formed by the weighted sum of all the historical global update directions as guidance for the training process. Therefore, to defend against the attacks, the key improvements are the selection of the reference direction $\mathbf{r}^t$ and the modified gradient $\mathbf{v}_m^t$.

\textbf{Reference direction:} 
% The primary improvement demanded here is the selection of the reference direction. Due to the malicious effects by the attackers, the reference direction formed by the weighted sum of all the historical global update directions will not suffice to serve as a guidance of the training process. 
The server is required to maintain a small root dataset $\mathcal{D}_{root}\in \mathcal{D}$ as in \cite{cao2020fltrust}. At each round $t$, the PS also updates a copy of the current global model $\boldsymbol{\theta}^t$ using the root dataset $\mathcal{D}_{root}$ for $U$ local iterations and arrives at the updated global model $\boldsymbol{\theta}^{t,U}$, i.e.,
\begin{equation}
\boldsymbol{\theta}^{t,u+1}=\boldsymbol{\theta}^{t,u}-\frac{\eta}{B} \sum_{b=1}^B \nabla F(\boldsymbol{\theta}^{t,u}; z_{b}^{t,u}), \label{update_root}
\end{equation}
for $u=0,...,U-1$, where we have $\boldsymbol{\theta}^{t,0}=\boldsymbol{\theta}^{t}$ and $z_b^{t,u}$ is drawn independently from dataset $\mathcal{D}_{root}$ across all batches, local iterations and training rounds.
The reference direction $\mathbf{r}^t$ is then set as 
\begin{align} \label{trust}
\mathbf{r}^t=:\boldsymbol{\theta}^{t,U}-\boldsymbol{\theta}^{t}.
\end{align}

\textbf{Vector manipulation:} Since the malicious client might scale the module of the local gradient, we can no longer use (\ref{vm}) as the vector manipulation formula because the module of the resulted $\mathbf{v}_m^t$ would be abnormally large or small. Instead, we normalize the module of each local gradient $\mathbf{g}_m^t$ according to the trusted reference direction $\mathbf{r}^t$ in (\ref{trust}) to defend against the module scaling attack, i.e., 
\begin{align}
    \mathbf{v}_m^t:=(1-\lambda_m^t)\frac{\|\mathbf{r}^t\|}{\|\mathbf{g}_m^t\|}\mathbf{g}_m^t+\lambda_m^t\mathbf{r}^t.\label{vm1}
\end{align}
The algorithm operates the same as the scenario without attacks (see Section IV B) with the above variables. 

\textbf{Remark 3.} 
% With this reference direction coming from a trusted source that provides a general update direction of the training process and the module normalization operation, the modified DRAG can easily tackle the byzantine attack mentioned in Section III. For the attack that scales the module of the local gradient, we tackle it by normalizing each gradient to avoid the module anomaly; for the attack that reverses the direction, (\ref{vm1}) can automatically reverse it back since $1-\lambda_m^t<0$ if the direction diverges too much from the reference direction (with the value of $c$ properly selected), as illustrated in Fig. 2.
With the reference direction obtained from a trusted source, which provides a general update direction for the training process, and the module normalization operation, the modified DRAG can effectively address the Byzantine attack mentioned in Section III. To counter the attack that scales the magnitude of the local gradient, we employ normalization on each gradient to prevent any anomaly in the magnitude. As for the attack that reverses the direction, with (\ref{vm1}), DRAG can automatically correct it by reversing the gradient back if $1-\lambda_m^t<0$, signifying excessive divergence from the reference direction (with the value of $c$ appropriately chosen), which is also illustrated in Fig. 2.

\section{Convergence Analysis of DRAG}
% The convergence proof of DRAG for non-convex objective functions, as in most machine learning tasks, is rigorously established in this subsection, before which we need two assumptions, as stated below.

In this section, we rigorously establish the convergence rate of DRAG for non-convex objective functions, which are prevalent in many machine learning tasks. We start by  introducing two assumptions that are utilized in the analysis, as stated below.

\textbf{Assumption 1} (Smoothness and Lower Boundedness) \textit{Each local 
function $ F_m(\boldsymbol{\theta})$ is $L$-smooth, i.e.,}
\begin{align}
    &\left\|\nabla F_m(\boldsymbol{\theta}_1)-\nabla F_m(\boldsymbol{\theta}_2)\right\| \leq L \left\|\boldsymbol{\theta}_1-\boldsymbol{\theta}_2\right\|,\label{lipschitz_gradient}
\end{align}
\textit{$\forall \boldsymbol{\theta}_1, \boldsymbol{\theta}_2\in \mathbb{R}^d$. The objective function $ F$ is also assumed to be lower-bounded by $ F^*$.}

Assumption 1 is the most common assumption used in convergence analysis, as in \cite{yang2020achieving}
\cite{yang2020achieving, zhou2018convergence,khaled2020tighter}.

\textbf{Assumption 2} (Unbiasedness and Bounded Variance) \textit{For the given model parameter $\boldsymbol{\theta}$, the local gradient estimator is unbiased, i.e.,}
\begin{align}
    \mathbb{E}[\nabla F_m(\boldsymbol{\theta};z)]=\nabla F_m(\boldsymbol{\theta}).
\end{align}
\textit{Moreover, both the variance of the local gradient estimator and that of the local gradient from the global one are bounded, i.e., there exist two constants $\sigma_L, \sigma_G>0$, such that}
\begin{align}
    &\mathbb{E}[\|\nabla F_m(\boldsymbol{\theta};z)-\nabla F_m(\boldsymbol{\theta})\|^2]\leq \sigma_L^2, \forall m\\
    &\mathbb{E}[\|\nabla F_m(\boldsymbol{\theta})-\nabla f(\boldsymbol{\theta})\|^2]\leq \sigma_G^2, \forall m.
\end{align}

Assumption 2 is also widely adopted when considering data heterogeneity, such as in \cite{yang2020achieving, haddadpour2019local, yu2019parallel}.

The above assumptions are sufficient to arrive at the following theorem, which is the upper bound for the expectation of the average squared gradient norm $\frac{1}{T}\mathbb{E}\left[\sum_{t=0}^{T-1}\left\|\nabla f(\boldsymbol{\theta}^t)\right\|^2\right]$.

\textbf{Theorem 1} \textit{Under Assumption 1 and 2, by choosing a steppsize $\eta$ satisfying $\eta\leq \frac{1}{8LU}$, there exists a positive constant $\gamma<\frac{1-3c}{2}-\left(5c+\frac{\eta LU}{2}\right)\left(90U^2L^2\eta^2+3\right)-15(1-c)L^2U^2\eta^2$, such that}
\begin{align}
    \frac{1}{T}\sum_{t=0}^{T-1}\|\nabla f(\boldsymbol{\theta}^{t})\|^2\leq \frac{f(\boldsymbol{\theta}^0)-f^*}{\gamma \eta U T} + V,
\end{align}
\textit{where} $V=\frac{1}{\gamma}\Bigg[\frac{4c\sigma_L^2}{BU}+\frac{\eta\sigma_L^2L}{2B}+\left(\frac{5c}{U}+\frac{\eta L}{2}\right)\big(15U^2L^2\eta^2V_1+3U\sigma_G^2)
+\frac{5(1-c)L^2\eta^2}{2}V_1\big)\Bigg]$ \textit{and} $V_1=\sigma_L^2+6U\sigma_G^2$.

Theorem 1 clearly states that provided the objective function is $L$-smooth
and the global and local variances of the gradients are bounded, the proposed DRAG can then finally converge at a sublinear speed. Further, by selecting a stepsize $\eta$ with the form $\mathcal{O}(1/\sqrt{T})$, we can readily achieve a convergence rate of the level $\mathcal{O}(1/\sqrt{T})$. 

% Note that the impact of the modifications of DRAG to the system setup can be detected in the expression of Theorem 1. As $c$ grows larger, the term $\frac{4c\sigma_L^2}{BU}+\frac{\eta\sigma_L^2L}{2B}+\left(\frac{5c}{U}+\frac{\eta L}{2}\right)\big(15U^2L^2\eta^2V_1+3U\sigma_G^2)$ in $V$ also increases, introducing a larger $V$; Meanwhile, as $c$ increases, the term $\frac{5(1-c)L^2\eta^2}{2}V_1$ decreases, which implies that the hyper-parameter $c$ has to be selected carefully to strike a balance.

\section{Simulation Results}

% In this section, we illustrate the advantages of DRAG by comparing its performance against state-of-the-art algorithms including SCAFFOLD \cite{karimireddy2020scaffold}, AdaBest \cite{varno2022adabest}, FedProx \cite{li2020federated}, and the vanilla FedAvg. These algorithms are tested on the EMNIST dataset and the more challenging CIFAR-10 dataset with partial and full worker participation to provide a more comprehensive perspective into the performances of these algorithms. We also on byzantine attack on these two datasets. 

In this section, we demonstrate the advantages of the DRAG algorithm by conducting a comprehensive performance comparison against state-of-the-art algorithms, including SCAFFOLD \cite{karimireddy2020scaffold}, AdaBest \cite{varno2022adabest}, FedProx \cite{li2020federated}, and the vanilla FedAvg \cite{mcmahan2017communication}. Our evaluation is carried out on both the EMNIST dataset and the CIFAR-10 dataset, with scenarios of both partial and full worker participation. Additionally, we assess the robustness of the algorithms against byzantine attacks on these two datasets. 

\subsection{Setting Up}
\textbf{EMNIST Dataset.} The EMNIST dataset \cite{cohen2017emnist} is an extended version of the famous MNIST dataset. Besides the handwritten digits from the MNIST dataset, the EMNIST dataset also includes handwritten letters. There are in total six different splits in the dataset and we used the ``balanced'' data split in this work with 47 balanced classes of data. Each piece of data is a $28\times 28$ grey image and there are 131,600 characters in the 47 balanced classes.

\textbf{CIFAR-10 Dataset.} The CIFAR-10 dataset \cite{krizhevsky2009learning} is composed of 60000 $32\times 32$ color images in a total of 10 classes,
with 6000 images in each class. There are 50000 images for training and 10000 for testing. Since it is a color image dataset, it is supposed to be more challenging than the EMNIST dataset.

\textbf{Byzantine Attack.} We set a total of $M=S=10$ clients and the dataset used here is the CIFAR-10 dataset. The byzantine attack initiated by the malicious client is as described in Section III. To be specific, the scalar $p_m^t$ conforms to normal distribution with zero mean and the variance is $\sigma^2=3$. The DRAG is also modified according to Section IV-C, with the root dataset maintaining $N_{root}=3000$ pieces of data samples drawn randomly from the global dataset.

\textbf{Data Heterogeneity.} As in \cite{cao2020fltrust}, we define the heterogeneity of data as follows. Any training data with label $\ell$ is assigned to client $\ell \bmod M$ with probability $q$ and to any other client with probability $\frac{1-q}{M-1}$. In dealing with client-drift, the high heterogeneous data distribution is implemented by setting $q=1$. In countering the byzantine attack, we define low data heterogeneity as $q=\frac{1}{M}$ and high data heterogeneity as $q=1$.

\subsection{Performance Analysis}

\begin{figure}[hbt]
\centering
\subfigure[full participation]{\includegraphics[width=3.5in]{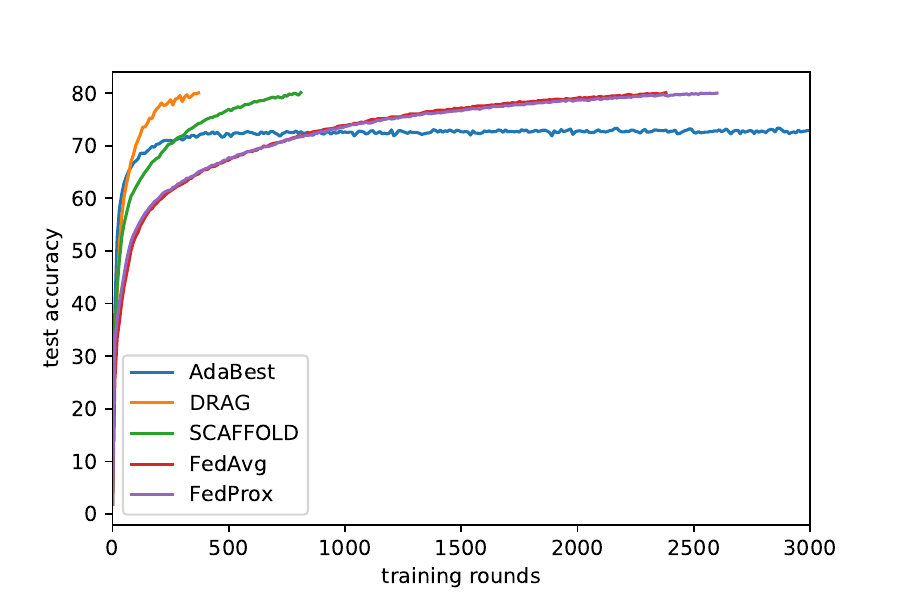}}

\subfigure[partial participation]{\includegraphics[width=3.5in]{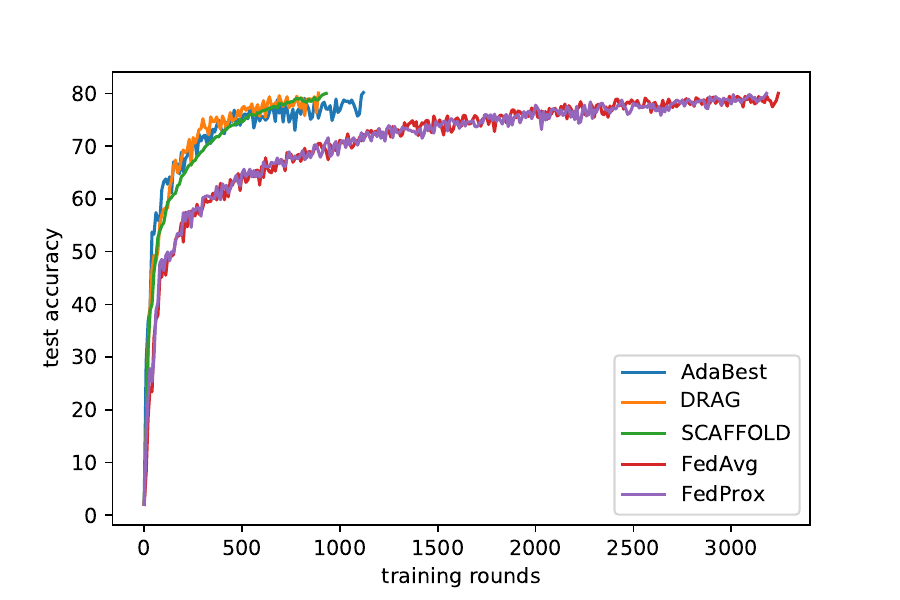}}
\caption{Performance comparison of DRAG with state-of-the-art algorithms under the EMNIST dataset with full and partial worker participation.}
\end{figure}

Fig. 3(a) and (b) plot the performance comparison of different algorithms under the EMNIST dataset ($q=1$), with full and partial worker participation, respectively. To be specific, there are in total $M=40$ workers and $S=10$ for partial worker participation. Each worker performs $U=5$ local updates. All the algorithms stop updating once the test accuracy reaches 80\%. The stepsize $\eta$ is 0.1. We used a two-layer fully connected network with 500 hidden units. The hyperparameters of the algorithms are listed below. For AdaBest, $\mu=0.02, \beta=0.8$; for DRAG, $c=0.25$, $\alpha=0.6$ for partial participation and $\alpha=1$ for full participation; $\mu=0.2$ for FedProx.

The comparison of different methods highlights the importance of addressing the client-drift issue. FedAvg and FedProx demonstrate subpar performance when neglecting this concern. While AdaBest is effective, it lacks robustness concerning the number of participating clients. Scaffold maintains relatively good performance in both cases, benefiting from its use of control variates to correct for client-drift in local updates. However, our proposed method, DRAG, exhibits significant superiority, especially in the presence of full participation. This advantage is attributed to the momentum-based reference direction, which provides a more accurate training direction, and the vector manipulation operation, which contributes to effective alignment with the reference direction.

% from Fig. 3 that AdaBest has the fastest convergence speed in the first stage with full participation but the final convergence accuracy is lower than the other algorithms. This is mainly because Adabest reduces the weight of historical information when utilizing control variates compared to SCAFFOLD. While SCAFFOLD outperforms AdaBest, FedAvg, and FedProx both under full and partial participation. However, DRAG is consistently the best for the two cases, thanks to the momentum-based reference direction that provides a relatively accurate training direction and the vector manipulation operation.

\begin{figure}[t!]
\centering
\subfigure[full participation]{\includegraphics[width=3.5in]{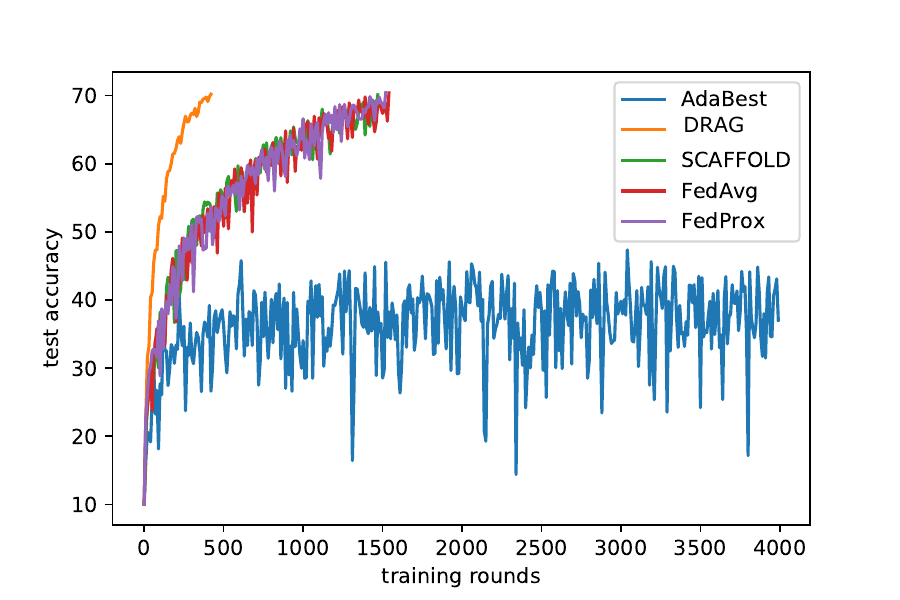}}

\subfigure[partial participation]{\includegraphics[width=3.5in]{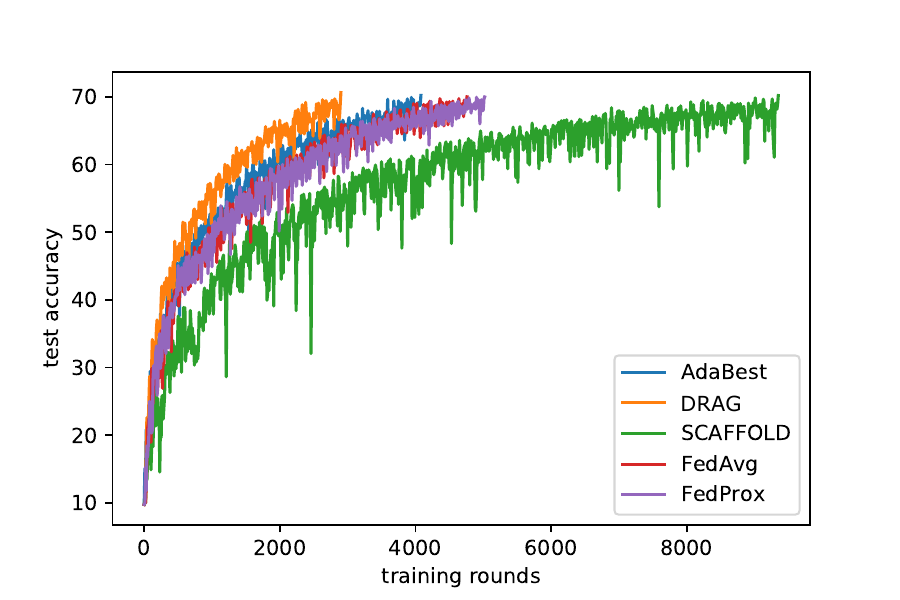}}
\caption{Performance comparison of DRAG with state-of-the-art algorithms under the CIFAR-10 dataset with full and partial worker participation.}
\end{figure}

In Fig. 4, we test the performances of these algorithms on the CIFAR-10 dataset ($q=1$) with $M=20$ workers for full participation and $S=5$ workers for partial participation. Each worker performs $U=5$ local updates. All the algorithms stop updating once the test accuracy reaches 70\%. The stepsize $\eta$ is 0.1. We used a CNN with two convolution layers and three fully-connected layers. For AdaBest, $\mu=0.02, \beta=0.2$; for DRAG, $c=0.1$, $\alpha=0.2$ for partial participation and $\alpha=1$ for full participation; $\mu=0.2$ for FedProx.

% In this setting, AdaBest still performs relatively better with partial participation; SCAFFOLD and FedProx do not have much improvement compared to FedAvg, as observed in \cite{li2021federated}. However, DRAG achieves distinctly better performance than other algorithms. To be specific, it only needs half of the training rounds to reach 70\% test accuracy compared to FedAvg with partial participation and the proportion becomes 1/4 in full participation.

In this setting, AdaBest still performs relatively better with partial participation; however, SCAFFOLD and FedProx show limited improvement compared to FedAvg, as previously observed in \cite{li2021federated}. On the other hand, DRAG outperforms all other algorithms significantly. Specifically, it achieves 70\% test accuracy with only half of the training rounds compared to FedAvg with partial participation, and this proportion further reduces to 1/4 under full participation.

% To sum up, with the proposed heuristic and intuitive design of algorithm instead of using control variates or regularization terms, DRAG is consistently the best algorithm to deal with client drift under various settings. It has strong adaptability thanks to the tunable hyper-parameters $c$ and $\alpha$, which are designed to balance the weights of the direction of the local gradient and the reference direction, and to develop a better reference direction, respectively.

To summarize, DRAG consistently outperforms other algorithms in handling client drift across various settings. Its success can be attributed to its heuristic and intuitive design, which avoids using control variates or regularization terms. Instead, DRAG leverages tunable hyper-parameters $c$ and $\alpha$ to balance the weights of the local gradient direction and the reference direction, respectively. This adaptability enables DRAG to develop a more accurate reference direction, further contributing to its superior performance.

\begin{figure}[t!]
\centering
\includegraphics[width=3.3in]{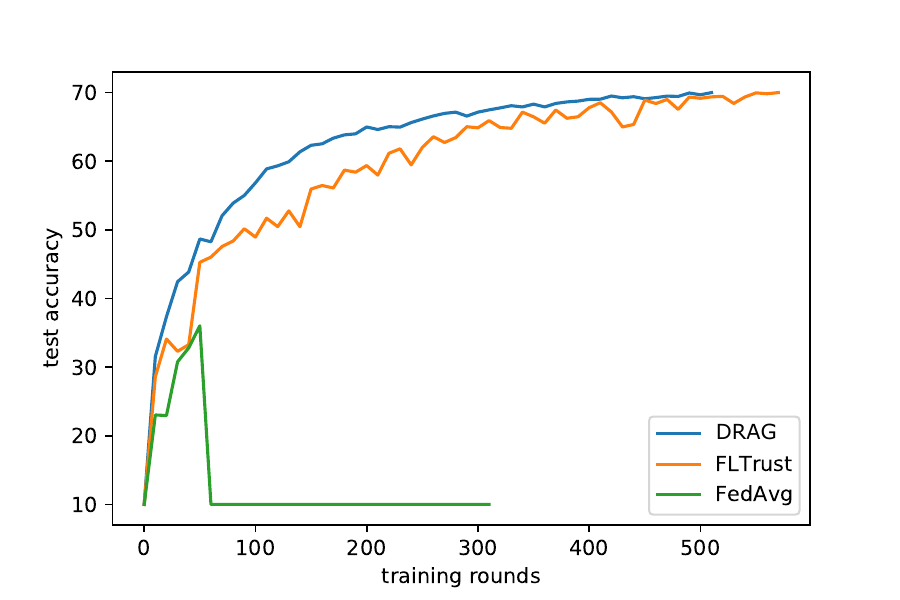}
\caption{Performance comparison of DRAG with state-of-the-art algorithms using the CIFAR-10 dataset under byzantine attacks with low data heterogeneity.} \label{byzan-iid}
\end{figure}

% Fig. 5 demonstrates the performances of DRAG, FLTrust \cite{cao2020fltrust} and FedAvg under byzantine attacks with low data heterogeneity. The algorithms stop training once the accuracy approaches 70\%. It is observed that the curve of FedAvg quickly falls down because it has no
% counter-attack techniques. And DRAG clearly achieves a better performance than FLTrust mainly because DRAG does not completely remove the attacker while taking advantage of its useful information through scaling and vector manipulation.

Fig.~\ref{byzan-iid} presents the performance of DRAG, FLTrust \cite{cao2020fltrust}, and FedAvg under byzantine attacks with low data heterogeneity ($q=\frac{1}{M}$), and there is $A=1$ attacker among the $M=S=10$ clients. In the plot, FedAvg's performance declines rapidly due to its lack of counter-attack techniques. In contrast, FLTrust demonstrates convergence by utilizing a small data sharing (root dataset) and ReLU-clipped cosine similarity. However, DRAG exhibits greater stability than FLTrust with just one data sharing step. The key advantage of DRAG lies in its ability to preserve the attacker's useful information through scaling and vector manipulation, instead of entirely removing the attacker. This property enables DRAG to achieve superior performance in handling byzantine attacks while maintaining the integrity of the FL process.

 % By leveraging the attacker's information, DRAG adapts dynamically to the presence of malicious clients, effectively mitigating their negative impact on the training process.

\begin{figure}[t!]
\centering
\includegraphics[width=3in]{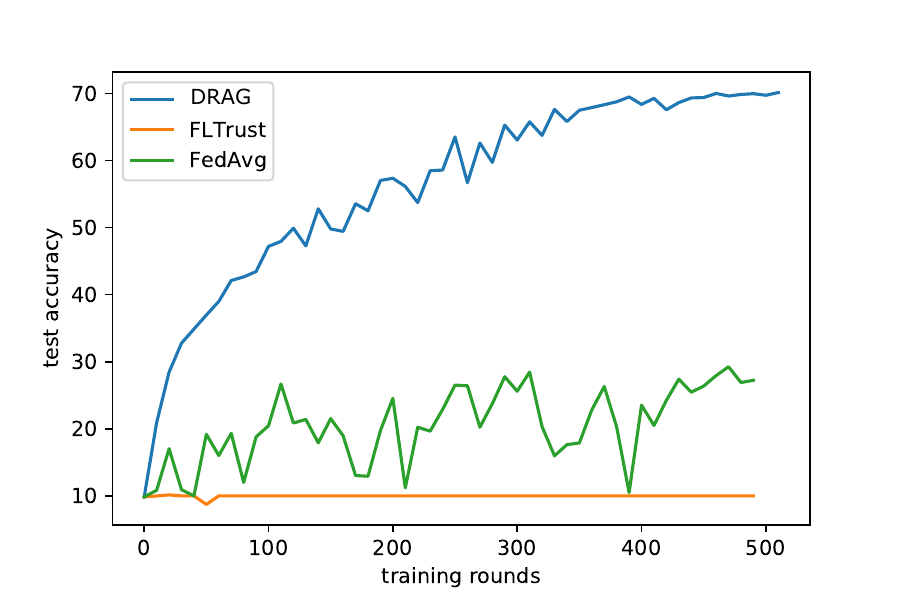}
\caption{Performance comparison of DRAG with state-of-the-art algorithms using the CIFAR-10 dataset under byzantine attacks with high data heterogeneity.} \label{byzan-noniid}
\end{figure}

In Fig.~\ref{byzan-noniid}, we increase the heterogeneity of data distribution among clients by setting $q=1$. Under this setting, the performance of FedAvg is still degraded as expected while FLTrust cannot converge at all. This is primarily due to the ReLU-clipping operation in FLTrust. As the heterogeneity of data increases, it is common that the local model update of a normal client diverges significantly from the trusted root direction. However, in FLTrust, such normal clients are erroneously identified as malicious and removed, causing the training process to crash. As analyzed above, DRAG retains its superiority even with the increased data heterogeneity.

% As previously analyzed, DRAG retains its superiority even with the increased data heterogeneity. Its advantageous properties, such as preserving useful information from attackers through scaling and vector manipulation instead of complete removal, allow DRAG to effectively handle byzantine attacks and maintain robustness in the face of highly heterogeneous data distributions.

\begin{figure}[t!]
\centering
\includegraphics[width=3.1in]{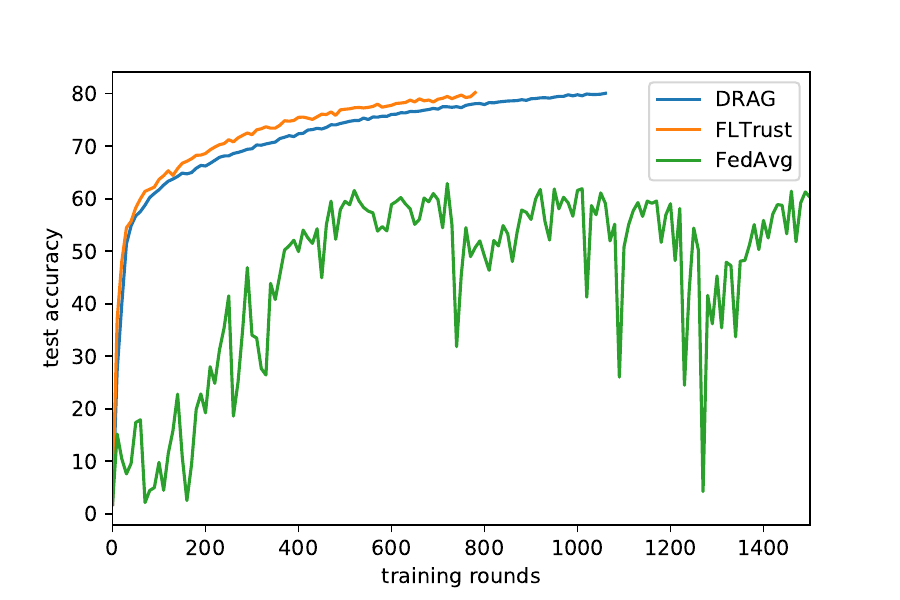}
\caption{Performance comparison of DRAG with state-of-the-art algorithms using the EMNIST dataset under byzantine attacks with low data heterogeneity.} \label{byzan-iidemnist}
\end{figure}

\begin{figure}[t!]
\centering
\includegraphics[width=3.3in]{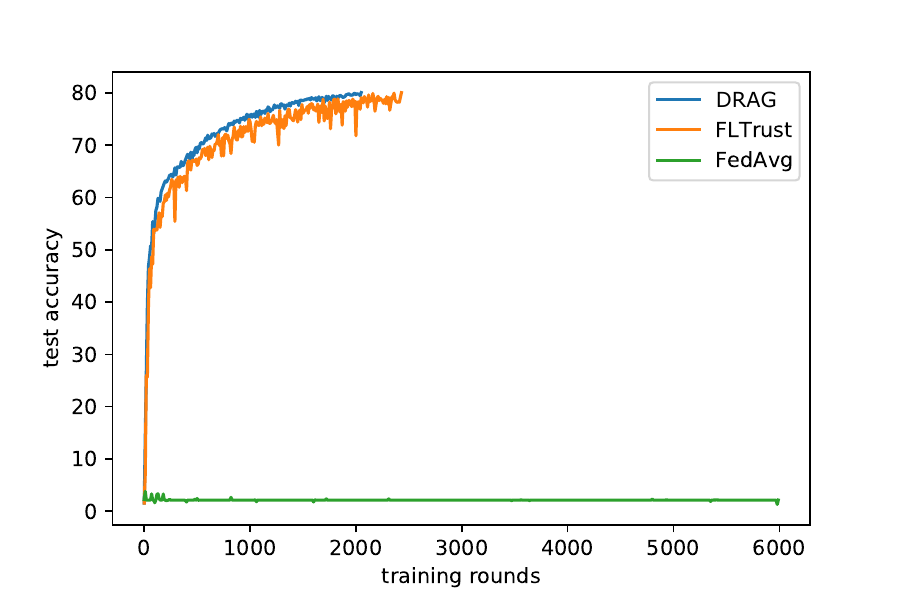}
\caption{Performance comparison of DRAG with state-of-the-art algorithms using the EMNIST dataset under byzantine attacks with high data heterogeneity.} \label{byzan-noniidemnist}
\end{figure}

To further substantiate the effectiveness of DRAG in handling byzantine attacks, we conduct performance testing on EMNIST, as shown in Fig.\ref{byzan-iidemnist} and Fig.\ref{byzan-noniidemnist}. In this setting, we increase the number of attackers to $A=4$ among a total of $M=10$ clients, as EMNIST is less challenging compared to CIFAR-10. We observe that FedAvg also achieves decent performance under low data heterogeneity since the malicious gradients are diluted through averaging, leading to a less negative impact on the training process compared to using CIFAR-10. Similarly, FLTrust performs well under low data heterogeneity, but the performance is limited when data is highly heterogeneously distributed, as it tends to mistake good clients for malicious ones. In contrast, DRAG consistently maintained its superiority across different data heterogeneity levels.

\section{Conclusion}
% In this work, we proposed a novel scheme justified by the name divergence-based adaptive aggregation (DRAG) to deal with the client drift issue in heterogeneous data distribution with local SGD. Unlike most methods that use control variates or regularization terms, the proposed DRAG algorithm heuristically drags each local gradient towards the reference direction according to its degree of divergence. The convergence analysis of DRAG is rigorously established to corroborate its feasibility theoretically. The performance of DRAG is also tested against state-of-the-art algorithms on EMNIST and CIFAR-10 datasets, the results of which substantiate the superiority of DRAG. Additionally, DRAG is proved to be robust to byzantine attacks that scale and reverse the direction of the local gradient.

% DRAG is strongly adaptive and has the potential in other scenarios such as its interaction with the wireless medium, its robustness against more complicated attacks, etc. These interesting directions will be explored in future works.

In this work, we introduce a novel scheme named divergence-based adaptive aggregation (DRAG) to address the client-drift issue in heterogeneous data distribution with local SGD. Unlike methods that rely on control variates or regularization terms, DRAG employs a heuristic approach, dynamically dragging each local gradient towards the reference direction based on the degree of divergence. We provide rigorous convergence analysis to theoretically support the feasibility of DRAG. Through extensive testing on EMNIST and CIFAR-10 datasets against state-of-the-art algorithms, DRAG demonstrates superior performance. Furthermore, we establish DRAG's resilience to byzantine attacks that scale and reverse the direction of the local gradient.

The adaptability of DRAG opens up exciting possibilities for exploring its utilization in other scenarios. To be specific, the wireless medium could be taken into consideration to explore more practical applications; we could also search for improvements of DRAG for attacks such as the label flipping attack, the Krum attack and so forth.

% To be specific, the wireless medium could be taken into consideration to explore more practical applications; we could also search for improvements of DRAG for attacks such as the label flipping attack, the Krum attack and so forth.

% \bibliography{bib}
% \bibliographystyle{iclr2022_conference}

%%%%%%%%%%%%%%%%%%%%%%%%%%%%%%%%%%%%%%%%%%%%%%%%%%%%%%%%%%%%

%%%%%%%%%%%%%%%%%%%%%%%%%%%%%%%%%%%%%%%%%%%%%%%%%%%%%%%%%%%%

\appendix

\section{Appendix}
% \subsection{Lemma 1}
% \textit{The global update $\Delta^t$ at round $t$ can be written in the form as}
% \begin{align}
%     \Delta^t=(1-c)\mathbf{g}^t+\frac{c}{M}\sum_{m\in\mathcal{M}}\frac{\left\langle\mathbf{g}_m^t, \mathbf{r}^t\right\rangle}{\|\mathbf{g}_m^t\|\|\mathbf{r}^t\|}\mathbf{g}_m^t+\frac{c}{M}\sum_{m\in\mathcal{M}}\frac{\|\mathbf{g}_m^t\|\|\mathbf{r}^t\|-\left\langle\mathbf{g}_m^t, \mathbf{r}^t\right\rangle}{\|\mathbf{r}^t\|^2}\mathbf{r}^t.\label{lemma1}
% \end{align}
% \begin{proof}
%     With the definition of $\Delta^t$, we can write:
%     \begin{align}
%         \Delta^t&=\frac{1}{M}\sum_{m=1}^M\left((1-\lambda_m^t)\mathbf{g}_m^t+\frac{\lambda_m^t\|\mathbf{g}_m^t\|}{\|\mathbf{r}^t\|}\mathbf{r}^t\right)\notag\\
%         \overset{()}{=}\frac{1}{M}\sum_{m=1}^M\left((1-\lambda_m^t)\mathbf{g}_m^t+\frac{\lambda_m^t\|\mathbf{g}_m^t\|}{\|\mathbf{r}^t\|}\mathbf{r}^t\right)\notag\\
%     \end{align}
% \end{proof}

\subsection{Proof of Theorem 1}
Due to the $L$-smoothness of the objective function, we have:
\begin{align}
    &\mathbb{E}[f(\boldsymbol{\theta}^{t+1})] \notag\\
    &\leq f(\boldsymbol{\theta}^{t}) + \left\langle\nabla f(\boldsymbol{\theta}^{t}), \mathbb{E}\left[\boldsymbol{\theta}^{t+1}-\boldsymbol{\theta}^t\right]\right\rangle + \frac{L}{2} \mathbb{E}\left[\|\boldsymbol{\theta}^{t+1}-\boldsymbol{\theta}^t\|^2\right]\notag\\
    &=f(\boldsymbol{\theta}^{t}) + \left\langle\nabla f(\boldsymbol{\theta}^{t}), \mathbb{E}\left[\Delta^t+a_1-a_1\right]\right\rangle + \frac{L}{2} \mathbb{E}\left[\|\Delta^t\|^2\right]\notag\\
       &=f(\boldsymbol{\theta}^{t}) -(1-c)\eta U\|\nabla f(\boldsymbol{\theta}^t)\|^2+ T_1+T_2.\label{one-step} 
    % &=f(\boldsymbol{\theta}^{t}) -(1-c)\eta U\|\nabla f(\boldsymbol{\theta}^t)\|^2+ \underbrace{\left\langle\nabla f(\boldsymbol{\theta}^{t}), \mathbb{E}\left[\Delta^t+a_1\right]\right\rangle}_{T_2} + \frac{L}{2} \underbrace{\mathbb{E}\left[\|\Delta^t\|^2\right]}_{T_1}.\label{one-step}
\end{align}
where we have defined the variable $a_1=(1-c)\eta U \nabla f(\boldsymbol{\theta}^t)$, $T_1=\mathbb{E}\left[\|\Delta^t\|^2\right]$, and $T_2=\left\langle\nabla f(\boldsymbol{\theta}^{t}), \mathbb{E}\left[\Delta^t+a_1\right]\right\rangle$. 

Next, the terms $T_1$ and $T_2$ will be bounded separately. First, with $T_1$ we have:
\begin{align}
    T_1&\overset{(a1)}{=}\mathbb{E}\left[\left\|\frac{1}{S}\sum_{m\in\mathcal{S}^t}\left((1-\lambda_m^t)\mathbf{g}_m^t+\frac{\lambda_m^t\|\mathbf{g}_m^t\|}{\|\mathbf{r}^t\|}\mathbf{r}^t\right)\right\|^2\right]\notag\\
    &\overset{(a2)}{\leq}\mathbb{E}\left[\left(\frac{1}{S}\sum_{m\in\mathcal{S}^t}\left((1-\lambda_m^t)\|\mathbf{g}_m^t\|+\frac{\lambda_m^t\|\mathbf{g}_m^t\|}{\|\mathbf{r}^t\|}\|\mathbf{r}^t\|\right)\right)^2\right]\notag\\
    &\overset{(a3)}{=}\frac{1}{S^2}\mathbb{E}\left[\left(\sum_{m\in\mathcal{S}^t}\|\mathbf{g}_m^t\|\right)^2\right]\notag\\
    &\overset{(a4)}{=}\frac{\eta^2}{S^2B^2}\left(\sum_{m\in\mathcal{S}^t}\mathbb{E}\left[\left\|\sum_{u=0}^{U-1}\sum_{b=1}^B\nabla F(\boldsymbol{\theta}_m^{t,u};z_{m,b}^{t,u})\right\|\right]\right)^2\notag\\
    &\overset{(a5)}{\leq}\frac{\eta^2}{SB^2}\sum_{m\in\mathcal{S}^t}\mathbb{E}\left[\left\|\sum_{u=0}^{U-1}\sum_{b=1}^B\nabla F(\boldsymbol{\theta}_m^{t,u};z_{m,b}^{t,u})\right\|^2\right]\notag\\
    &\overset{(a6)}{=}\frac{\eta^2}{SB^2}\sum_{m\in\mathcal{S}^t}\mathbb{E}\left[\left\|\sum_{u=0}^{U-1}\sum_{b=1}^Ba_2\right\|^2\right]\nonumber \\
   & ~~~~~~+\frac{\eta^2}{S}\sum_{m\in\mathcal{S}^t}\mathbb{E}\left[\left\|\sum_{u=0}^{U-1}\nabla F_m(\boldsymbol{\theta}_m^{t,u})\right\|^2\right]\notag\\
    &\overset{(a7)}{=}\frac{\eta^2U\sigma_L^2}{B}+\frac{\eta^2}{M}\sum_{m\in\mathcal{M}}\mathbb{E}\left[\left\|\sum_{u=0}^{U-1}\nabla F_m(\boldsymbol{\theta}_m^{t,u})\right\|^2\right]
    % &\overset{(a8)}{=}\frac{\eta^2U\sigma_L^2}{B}+\frac{\eta^2}{M}\sum_{m\in\mathcal{M}}\mathbb{E}\left[\left\|\sum_{u=0}^{U-1}\nabla F_m(\boldsymbol{\theta}_m^{t,u})\right\|^2\right]\notag\\ 
    \label{T1}
\end{align}
where $(a1)$ is due to the definition of $\Delta^t$; $(a2)$ comes from triangle inequality, i.e., $\|\mathbf{a}+\mathbf{b}\|\leq \|\mathbf{a}\|+\|\mathbf{b\|}$; $(a3)$ comes from direct computation; $(a4)$ is due to the definition of $\mathbf{g}_m^t$; $(a5)$ is due to Cauchy-Schwartz inequality; $(a6)$ is because of $\mathbb{E}[\|x\|^2]=\mathbb{E}[\|x-\mathbb{E}[x]\|^2+\|\mathbb{E}[x]\|^2]$, $\mathbb{E}[\nabla F_m(\boldsymbol{\theta}_{m}^{j,u};z_{m,b}^{j,u})]=\nabla F_m(\boldsymbol{\theta}_{m}^{j,u})$ and the definition of variable $a_2=\nabla F_m(\boldsymbol{\theta}_m^{t,u};z_{m,b}^{t,u})-\nabla F_m(\boldsymbol{\theta}_m^{t,u})$, and $(a7)$ is due to the fact that $\mathbb{E}[\|x_1+...+x_n\|^2]=\mathbb{E}[\|x_1\|^2+...+\|x_n\|^2]$ if $x_i'$s are independent with zero mean, together with the fact that the probability of each worker being selected without replacement is $\frac{S}{M}$.

For the term $T_2$, we have:
\begin{align}
    T_2&\overset{(b1)}{=}\Bigg\langle\nabla f(\boldsymbol{\theta}^{t}), \mathbb{E}\Bigg[(1-c)\mathbf{g}^t+\frac{c}{S}\sum_{m\in\mathcal{S}^t}\frac{\left\langle\mathbf{g}_m^t, \mathbf{r}^t\right\rangle}{\|\mathbf{g}_m^t\|\|\mathbf{r}^t\|}\mathbf{g}_m^t\nonumber\\
    &+\frac{c}{S}\sum_{m\in\mathcal{S}^t}\frac{\|\mathbf{g}_m^t\|\|\mathbf{r}^t\|-\left\langle\mathbf{g}_m^t, \mathbf{r}^t\right\rangle}{\|\mathbf{r}^t\|^2}\mathbf{r}^t+(1-c)\eta U \nabla f(\boldsymbol{\theta}^t)\Bigg]\Bigg\rangle\notag\\
    &\overset{(b2)}{=}\underbrace{\left\langle\nabla f(\boldsymbol{\theta}^{t}), \mathbb{E}\left[(1-c)\mathbf{g}^t+a_1\right]\right\rangle}_{T_{2,1}}\nonumber\\
    &+\underbrace{\left\langle\nabla f(\boldsymbol{\theta}^{t}), \mathbb{E}\left[\frac{c}{M}\sum_{m\in\mathcal{S}^t}\frac{\left\langle\mathbf{g}_m^t, \mathbf{r}^t\right\rangle}{\|\mathbf{g}_m^t\|\|\mathbf{r}^t\|}\mathbf{g}_m^t\right]\right\rangle}_{T_{2,2}}\notag\\
    &+\underbrace{\left\langle\nabla f(\boldsymbol{\theta}^t), \mathbb{E}\left[\frac{c}{M}\sum_{m\in\mathcal{S}^t}\frac{\|\mathbf{g}_m^t\|\|\mathbf{r}^t\|-\left\langle\mathbf{g}_m^t, \mathbf{r}^t\right\rangle}{\|\mathbf{r}^t\|^2}\mathbf{r}^t\right]\right\rangle}_{T_{2,3}}, \label{T2}
\end{align}
where $(b1)$ comes from the definition of $\Delta^t$ and $(b2)$ comes from decomposition, where $\mathbf{g}^t=\frac{1}{S}\sum_{m\in\mathcal{S}^t}\mathbf{g}_m^t$. The three terms $T_{2,1}$, $T_{2,2}$ and $T_{2,3}$ are then bounded separately.

For the term $T_{2,1}$, we have:
\begin{align}
    &T_{2,1} = \left\langle\nabla f(\boldsymbol{\theta}^{t}), \mathbb{E}\left[(1-c)\mathbf{g}^t+a_1\right]\right\rangle\notag\\
    &\overset{(c1)}{=} \left\langle\nabla f(\boldsymbol{\theta}^{t}), \mathbb{E}\left[(1-c)\bar{\mathbf{g}}^t+a_1\right]\right\rangle\notag\\
    &\overset{(c2)}{=}\Bigg\langle\nabla f(\boldsymbol{\theta}^{t}), \mathbb{E}\Bigg[-(1-c)\frac{1}{M}\sum_{m=1}^M\sum_{u=0}^{U-1}\eta\nabla F_m(\boldsymbol{\theta}_m^{t,u})\nonumber\\
    &+(1-c)\eta U\frac{1}{M}\sum_{m=1}^M \nabla F_m(\boldsymbol{\theta}^t)\Bigg]\Bigg\rangle\notag\\
    &\overset{(c3)}{=}\Bigg\langle\sqrt{\eta U(1-c)}\nabla f(\boldsymbol{\theta}^{t}),-\frac{\sqrt{\eta(1-c)}}{M\sqrt{U}}\mathbb{E}\left[\sum_{m=1}^M\sum_{u=0}^{U-1}a_3\right]\Bigg\rangle\notag\\
    &\overset{(c4)}{\leq}a_4+\frac{(1-c)\eta}{2UM^2}\mathbb{E}\left[\left\|\sum_{m=1}^M\sum_{u=0}^{U-1}a_3\right\|^2\right]\notag\\
    &\overset{(c5)}{\leq}a_4+\frac{(1-c)\eta}{2M}\sum_{m=1}^M\sum_{u=0}^{U-1}\mathbb{E}\left[\left\|a_3\right\|^2\right]\notag\\
    &\overset{(c6)}{\leq}a_4+\frac{(1-c)\eta L^2}{2M}\sum_{m=1}^M\sum_{u=0}^{U-1}\mathbb{E}\left[\left\|\boldsymbol{\theta}_m^{t,u}-\boldsymbol{\theta}^t\right\|^2\right]
    % -\frac{(1-c)\eta}{2UM^2}\mathbb{E}\left[\left\|\sum_{m=1}^M\sum_{u=0}^{U-1}\nabla F_m(\boldsymbol{\theta}_m^{t,u})\right\|^2\right]
    ,\label{T21}
\end{align}
where $(c1)$ is due to the fact that the sampling distribution is identical at every round; $(c2)$ comes from the definition of $\Bar{\mathbf{g}}^t=\frac{1}{M}\sum_{m\in\mathcal{M}}\mathbf{g}_m^t$ and $f(\boldsymbol{\theta}^t)$; $(c3)$ comes from direct computation and the defined variable $a_3=\nabla F_m(\boldsymbol{\theta}_m^{t,u})-\nabla F_m(\boldsymbol{\theta}^t)$; $(c4)$ follows from $\langle\mathbf{x},\mathbf{y}\rangle\leq\frac{1}{2}[\|\mathbf{x}\|^2+\|\mathbf{y}\|^2]$ and the defined variable $a_4=\frac{(1-c)\eta U}{2}\|\nabla f(\boldsymbol{\theta}^{t})\|^2$; $(c5)$ uses Cauchy-Schwartz inequality; and $(c6)$ is due to the $L$-smoothness assumption.

Similarly, by following the bounding operation of $T_{2,1}$, we can have the following two inequalities
\begin{align}
    T_{2,2}\leq a_4+\frac{c\eta}{SU}\sum_{m\in\mathcal{S}^t}\mathbb{E}\left[\left\|\sum_{u=0}^{U-1}\nabla F_m(\boldsymbol{\theta}_m^{t,u})\right\|^2\right],\label{T22}
\end{align}

\begin{align}
    T_{2,3}\leq a_4+\frac{4c\eta\sigma_L^2}{B}+\frac{4c\eta}{MU}\sum_{m\in\mathcal{M}}\mathbb{E}\left[\left\|\sum_{u=0}^{U-1}\nabla F_m(\boldsymbol{\theta}_m^{t,u})\right\|^2\right],\label{T23}
\end{align}

We continue to bound the following term:
\begin{align}
    &\mathbb{E}\left[\left\| \sum_{u=0}^{U-1}\nabla F_m(\boldsymbol{\theta}_{m}^{t,u})\right\|^2\right] \notag\\
    &=\mathbb{E}\Bigg[\Bigg\| \sum_{u=0}^{U-1}(a_3 +\nabla F_m(\boldsymbol{\theta}^t)-\nabla f(\boldsymbol{\theta}^t)+\nabla f(\boldsymbol{\theta}^t))\Bigg\|^2\Bigg]\notag\\    
    % &=\mathbb{E}\Bigg[\Bigg\| \sum_{u=0}^{U-1}(\nabla F_m(\boldsymbol{\theta}_{m}^{t,u})-\nabla F_m(\boldsymbol{\theta}^t) +\nabla F_m(\boldsymbol{\theta}^t)\notag\\
    % &-\nabla f(\boldsymbol{\theta}^t)+\nabla f(\boldsymbol{\theta}^t))\Bigg\|^2\Bigg]\notag\\
    &\overset{(f1)}{\leq} 3UL^2\sum_{u=0}^{U-1}\mathbb{E}[\|\boldsymbol{\theta}_{m}^{t,u}-\boldsymbol{\theta}^t\|^2]+3U^2\sigma_G^2+3U^2\|\nabla f(\boldsymbol{\theta}^t)\|^2\notag\\
    % &\overset{(f2)}{\leq} 15U^3L^2\eta^2(\sigma_L^2+6U\sigma_G^2)+(90U^4L^2\eta^2+3U^2)\|\nabla f(\boldsymbol{\theta}^t)\|^2+3U^2\sigma_G^2\notag\\
    &\overset{(f2)}{=}C_1\|\nabla f(\boldsymbol{\theta}^t)\|^2+C_2,\label{ti}
\end{align}
where $(f1)$ is due to Cauchy-Schwartz inequality; $(f2)$ is from \cite[Lemma 3]{reddi2020adaptive}, which proves the inequality 
\begin{align}
\mathbb{E}\left[\left\|\boldsymbol{\theta}_{m}^{t,u}-\boldsymbol{\theta}^t\right\|^2\right]\!\leq\! 5U\eta^2(\sigma_L^2 \!+ \! 6U\sigma_G^2) \!+ \! 30U^2\eta^2 \left\|\nabla f(\boldsymbol{\theta}^t)\right\|^2;\notag 
\end{align}
and we have the definitions with $C_1=90U^4L^2\eta^2+3U^2$ and $C_2=15U^3L^2\eta^2(\sigma_L^2+6U\sigma_G^2)+3U^2\sigma_G^2$.

% Substituting (\ref{T21}), (\ref{T22}) and (\ref{T23}) into (\ref{T2}), we have:
% \begin{align}
%     T_2 &= \frac{(1+c)\eta U}{2}\|\nabla f(\boldsymbol{\theta}^{t})\|^2+\frac{4c\eta\sigma_L^2}{B}\notag\\
%     &+\frac{5c\eta}{U}(C_1\|\nabla f(\boldsymbol{\theta}^t)\|^2+C_2)\notag\\
%     &\!\!+\!\frac{(1\!-\!c)\eta L^2U}{2}(5U\eta^2(\sigma_L^2+6U\sigma_G^2)+30U^2\eta^2\|\nabla f(\boldsymbol{\theta}^{t})\|^2).\label{T_2final}
% \end{align}

Hence, we can bound $T_2$ by using (\ref{T21}), (\ref{T22}) and (\ref{T23}). Together with (\ref{T1}), we have 
% Substituting (\ref{T_2final}) and (\ref{T1}) into (\ref{one-step}), we have:
% \begin{align}
%     &\mathbb{E}[f(\boldsymbol{\theta}^{t+1})]\leq f(\boldsymbol{\theta}^{t})-(1-c)\eta U\|\nabla f(\boldsymbol{\theta}^{t})\|^2\notag\\
%     &+ \frac{(1\!+\!c)\eta U}{2}\|\nabla f(\boldsymbol{\theta}^{t})\|^2\!+\!\frac{4c\eta\sigma_L^2}{B}+\frac{5c\eta}{U}(C_1\|\nabla f(\boldsymbol{\theta}^t)\|^2\!+\!C_2)\notag\\
%     &+\frac{(1-c)\eta L^2U}{2}(5U\eta^2(\sigma_L^2+6U\sigma_G^2)+30U^2\eta^2\|\nabla f(\boldsymbol{\theta}^{t})\|^2) \notag\\
%     &+\frac{\eta^2U\sigma_L^2L}{2B}+\frac{\eta^2L}{2}(C_1\|\nabla f(\boldsymbol{\theta}^t)\|^2+C_2)\notag\\
%     &=f(\boldsymbol{\theta}^{t})-\eta U\Bigg(\frac{1-3c}{2}-\left(\frac{5c}{U}+\frac{\eta L}{2}\right)\left(90U^3L^2\eta^2+3U\right)\notag\\
%     &-\frac{30(1-c)L^2U^2\eta^2}{2}\Bigg)\|\nabla f(\boldsymbol{\theta}^{t})\|^2+C_3,\label{onestepnew}
% \end{align}
\begin{align}
    &\mathbb{E}[f(\boldsymbol{\theta}^{t+1})] \notag\\
    &\leq f(\boldsymbol{\theta}^{t})-\eta U\Bigg(\frac{1-3c}{2}-\left(\frac{5c}{U}+\frac{\eta L}{2}\right)\left(90U^3L^2\eta^2+3U\right)\notag\\
    &-\frac{30(1-c)L^2U^2\eta^2}{2}\Bigg)\|\nabla f(\boldsymbol{\theta}^{t})\|^2+C_3,\label{onestepnew}
\end{align}
where $C_3=\frac{4c\eta\sigma_L^2}{B}+\frac{\eta^2U\sigma_L^2L}{2B}+(\frac{5c\eta}{U}+\frac{\eta^2L}{2})C_2+\frac{5(1-c)\eta^3 L^2U^2}{2}(\sigma_L^2+6U\sigma_G^2)$.

Rearranging and summing from $t=0,...,T-1$, we have:
\begin{align}
    \frac{1}{T}\sum_{t=0}^{T-1}\|\nabla f(\boldsymbol{\theta}^{t})\|^2\leq \frac{f(\boldsymbol{\theta}^0)-f^*}{\gamma \eta U T} + V,
\end{align}
where there exists a constant $\gamma$ satisfying $\frac{1-3c}{2}-\left(\frac{5c}{U^2}+\frac{\eta L}{2U}\right)C_1-\frac{30(1-c)L^2U^2\eta^2}{2}>\gamma>0$ and 
$V=\frac{C_3}{\gamma U\eta}$. 
% $V=\frac{1}{\gamma}\left[\frac{4c\sigma_L^2}{BU}+\frac{\eta\sigma_L^2L}{2B}+\left(\frac{5c}{U^2}+\frac{\eta L}{2U}\right)C_2+\frac{5(1-c)L^2U\eta^2}{2}(\sigma_L^2+6U\sigma_G^2)\right]$. 
The proof is then complete.

\bibliographystyle{IEEEtran}

\end{document}